\documentclass{article}
\usepackage{iclr2027_conference,times}
\usepackage{hyperref}
\usepackage{url}
\usepackage{float}
\usepackage{booktabs}
\usepackage[table]{xcolor}
\usepackage{xspace}
\usepackage{xcolor}
\usepackage{graphicx}
\usepackage{amsmath}
\usepackage{algorithm}
\usepackage{algpseudocode}
\usepackage{wrapfig}
\usepackage{subcaption}
\usepackage{listings}
\usepackage{array}
\usepackage[skins]{tcolorbox}

\definecolor{sectionbg}{RGB}{245,240,235}
\definecolor{highlightbg}{RGB}{232,233,246}
\definecolor{recentcolor}{HTML}{4477AA}
\definecolor{globalcolor}{HTML}{EE7733}
\definecolor{eventcolor}{HTML}{228833}
\definecolor{controlcolor}{HTML}{AA3377}
\newcommand{\sys}{MemEvo\xspace}

\newcommand{\etc}{etc.\xspace}

\newcommand{\figsubref}[2]{%
  \hyperref[#2]{\ref*{#1}\subref*{#2}}%
}
\newcommand{\br}{\\[2mm]}
\lstdefinestyle{memorydsl}{
    basicstyle=\ttfamily\footnotesize,
    columns=fullflexible,
    keepspaces=true,
    showstringspaces=false,
    frame=none,
    escapeinside={(*@}{@*)},
}
\newcolumntype{L}[1]{>{\raggedright\arraybackslash}m{#1}}
\newcolumntype{C}[1]{>{\centering\arraybackslash}m{#1}}
\newcolumntype{R}[1]{>{\raggedleft\arraybackslash}m{#1}}

\title{\sys: Automatic Discovery of Streaming Video Memory Mechanisms}
\arxivcopy

\author{Guohong Liu$^{1}$
\qquad Jialei Ye$^{2}$
\qquad Shanhui Zhao$^{1}$
\qquad Yunxin Liu$^{1}$
\qquad Yuanchun Li$^{1,\dagger}$\\[2mm]
$^{1}$Institute for AI Industry Research, Tsinghua University\\
$^{2}$Peking University\\
$^{\dagger}$Corresponding Author (liyuanchun@air.tsinghua.edu.cn)}

\begin{document}
\maketitle

\begin{abstract}
Query-agnostic streaming video understanding requires vision-language models to continuously 
compress an indefinitely growing visual stream into a bounded memory before future queries are known. 
The performance depends critically on the memory mechanism—what observations to preserve, 
how to represent and consolidate them, and what information to retrieve when a query eventually arrives. 
Rather than designing a single memory architecture by hand, we formulate memory design as a search problem 
over executable memory programs. We introduce a lightweight domain-specific language that expresses memory 
mechanisms through structured primitives for representation, admission, retention, consolidation, budgeting, 
and retrieval, while enforcing causal and bounded-memory constraints. Although structured, the derived 
program space remains large and contains heterogeneous, conditionally dependent design choices whose effects 
can only be assessed via downstream execution. We therefore propose \sys, an LLM-driven auto-research 
framework that uses pretrained LLM as a semantics-aware proposal model to iteratively generate and refine 
candidate memory programs based on accumulated experimental feedback. At runtime, a deterministic evaluation pipeline 
validates and evaluates each candidate, while the underlying vision-language model remains frozen throughout 
discovery. We finally produce a training-free, bounded-memory mechanism. 
Extensive experiments on StreamingBench and OVO-Bench demonstrate strong 
streaming video understanding performance together with substantial context and inference efficiency.
\end{abstract}

\section{Introduction}
Vision language models (VLMs) have extended visual-language perception ability from images to videos~
\citep{Flamingo,BLIP-2,LLaVA,VideoLLM-online,LLaVA-Video,qwen3-vl,InternVL3}.
A natural next step is to move beyond reasoning over completed video clips to continuously operating visual systems.
However, the model cannot wait until a video ends in such settings: it must process an open-ended
stream as observations arrive and remain ready to answer queries at arbitrary future times.
This creates a fundamental memory problem. An indefinitely growing visual history must be continuously
transformed into a bounded state, and for the query-agnostic cases considered in this work, the memory
decisions must be made \emph{before future queries are known}. Information discarded early may turn
out to be essential, while indiscriminately retaining history could be computationally infeasible.
So the capability of a streaming video system depends not only on the underlying MLLM, but critically
on the memory mechanism that determines what information is preserved, how it is compressed, and when it is forgotten.

Existing approaches instantiate such memory mechanisms in different ways.
Training-based methods introduce streaming-specific objectives, data, or architectures to adapt models
to continuous visual inputs~\citep{VideoLLM-online,StreamForest}, whereas training-free ones modify
inference-time memory of frozen models through KV-cache reuse, visual-token compression, retrieval, or
hierarchical memory organization~\citep{ReKV,StreamBridge,InfiniPot-V,HERMES,FluxMem,OASIS}. These all
demonstrate that memory design substantially affects long-range retention, inference efficiency and
downstream understanding. Yet most existing methods ultimately commit to a particular manually designed mechanism.
Designing such a mechanism is itself a nontrivial algorithmic problem. The memory must jointly specify
\emph{what} observations to admit, \emph{how} they should be represented, \emph{when} they should be retained,
discarded, or merged, \emph{how} limited capacity should be allocated across different forms of memory, and
\emph{what} information should be retrieved once a query arrives. These are heterogeneous and
strongly coupled.
Moreover, their downstream effect is generally only revealed after executing the complete streaming
and answering pipeline. Such structural choices are discrete and largely non-differentiable, and
therefore cannot be optimized as ordinary model parameters through gradient descent.

\begin{figure}[H]
    \centering
    \includegraphics[width=0.85\textwidth]{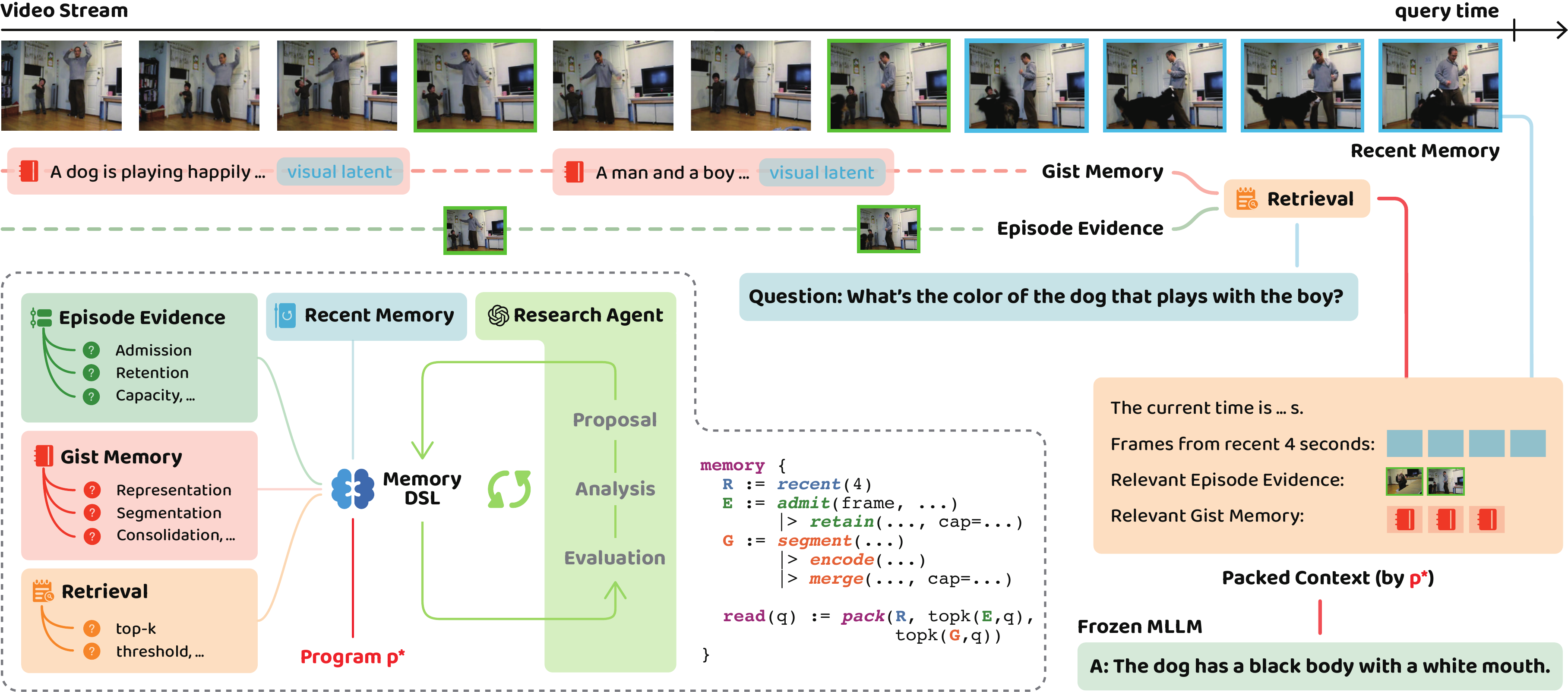}
    \caption{Overview of \sys. Streaming memory mechanism is represented as a structured
    program over three functional memory roles: recent memory, episodic evidence, and gist memory. 
    During the offline discovery stage, an LLM-based research agent proposes candidate 
    programs based on accumulated experimental feedback. 
    At deployment, the selected memory program is executed to causally maintain bounded memory and construct 
    query-time context for the frozen MLLM.}
    \label{fig:teaser}
\end{figure}

We take the view that a streaming memory mechanism can instead be represented as an \emph{executable memory program}.
Such a program specifies how incoming observations update a bounded memory state and how it is read
when a future query arrives. We introduce a lightweight domain-specific program space
composed of structured memory primitives. This makes the object
being optimized explicit: rather than directly optimizing the contents of memory for individual videos,
we search for the \emph{program that maintains better memory} across streams.
Though this representation structures the design problem, the resulting search space
remains large. Critically, it is not merely a Cartesian product of interchangeable hyperparameters:
many choices have algorithmic semantics, conditional dependencies, and interactions that determine which
modifications are meaningful. Exhaustively evaluation is impractical, while blindly
perturbing parameters ignores useful knowledge about the functions of different memory operations.
Recent LLM-based research agents have demonstrated an ability to iteratively generate hypotheses,
implement or revise candidate solutions, and incorporate experimental feedback when exploring complex
algorithmic spaces~\citep{ResearchAgent,Agent-Lab,AIScientist}. This motivates us to
introduce \textbf{\sys}, an LLM-driven framework for automatic discovery of streaming video memory
mechanisms. The LLM acts as a semantics-aware proposal model: given the available memory primitives
and the history of previous experiments, it proposes explicit candidate programs and refinements.
A fixed evaluation pipeline then checks their validity, executes the complete streaming system, and
measures task quality and computational cost. Experimental outcomes, rather than the LLM itself,
determine which candidates are retained and used to guide subsequent exploration, as depicted in Figure~\ref{fig:teaser}.

To make this search tractable and interpretable, we organize memory around three complementary functional
roles. \emph{Recent memory} preserves high-fidelity observations from the immediate past and serves
as a fixed short-term perceptual anchor. \emph{Episodic evidence} stores sparse, high-fidelity observations
from earlier moments that may provide precise historical evidence. \emph{Gist memory} compresses longer
temporal ranges into lower-cost representations that retain broader semantic context. Within this
factorization, \sys searches coupled program choices over representation, admission, retention, segmentation,
consolidation, budget allocation, and query-time retrieval. The underlying MLLM remains frozen throughout
discovery; only the memory program changes.

Extensive experiments show that the memory program discovered by \sys transfers effectively across
streaming video benchmarks and MLLM backbones. On StreamingBench and OVO-Bench, it substantially
improves the corresponding frozen backbone while maintaining bounded memory. Fine-grained ablations
further show that recent, episodic, and gist memory provide complementary benefits across different
types of questions. Resource profiling demonstrates that the discovered mechanism substantially
reduces reader context length and end-to-end inference latency compared with directly processing
uniformly sampled video history.

Our contributions are threefold:
\begin{itemize}
    \item We formulate query-agnostic streaming video memory design as a search problem over 
    \emph{executable memory programs}, and define a structured program space with explicit causal 
    and bounded constraints over complementary recent, episodic, and gist memory functions.

    \item We propose \sys, an LLM-driven memory discovery framework where a pretrained LLM uses 
    program semantics and accumulated experimental feedback to iteratively propose and refine 
    candidate memory mechanisms, while a deterministic evaluation pipeline validates their quality and cost.

    \item We extensively evaluate the discovered memory across benchmarks 
    and MLLMs, demonstrating substantial improvements over frozen backbones, complementary effects 
    among different memory components, and favorable context and inference efficiency.
\end{itemize}

\section{Related Work}
\paragraph{Memory for Long and Streaming Video Understanding.}
Memory is widely used to extend multimodal models beyond short, pre-segmented videos. Early long-video 
systems maintain bounded short-/long-term or compressed memory banks~\citep{MovieChat,MA-LMM}, while 
streaming-oriented models further introduce compact contextual memory, persistent event memory, or 
temporally aligned processing~\citep{Flash-VStream,VideoLLM-online,StreamBridge,StreamForest}. Recent 
training-free approaches instead modify the inference-time memory of frozen MLLMs through KV-cache 
retrieval or compression~\citep{ReKV,InfiniPot-V,HERMES}, adaptive visual-token consolidation~\citep{FluxMem}, 
and hierarchical event memory with retrieval~\citep{OASIS}. Other concurrent methods explore semantic-aware 
or visually grounded retention mechanisms~\citep{SAVEMem,CurveStream,FOLIO}. Collectively, these works 
establish memory design as a key factor in streaming video understanding, but primarily instantiate manually 
specified mechanisms. In contrast, we treat the memory mechanism itself as the object of discovery, exposing 
representation, admission, retention, consolidation, budgeting, and retrieval as a structured program space 
to be searched automatically.

\paragraph{LLM-based Automated Research and Algorithm Discovery.}
LLM agents have increasingly automated research workflows involving hypothesis generation, experiment 
execution, analysis, and iterative refinement~\citep{ResearchAgent,MLR-Copilot,Agent-Lab,AIScientist,AIScientist-v2}.
A complementary line of work uses LLMs to directly search over executable algorithms and systems. EoH combines 
LLM-generated heuristic ideas and code with evolutionary search~\citep{EoH}; ADAS uses a meta-agent to discover 
agentic systems represented as programs~\citep{ADAS}; and AlphaEvolve iteratively modifies algorithmic code using 
feedback from automated evaluators~\citep{AlphaEvolve}. These studies demonstrate the potential of LLMs as proposal 
mechanisms for exploring structured algorithmic spaces under empirical feedback. \sys applies this paradigm to 
streaming video memory: rather than generating a research artifact or optimizing a general-purpose algorithm, 
it searches a domain-specific space of executable memory programs, whose candidates are validated and evaluated 
through real downstream experiments while the underlying MLLM remains frozen.

\section{Method}
\label{sec:method}
We cast streaming memory design as search over executable programs in a compact domain-specific 
language (DSL). The DSL fixes the high-level memory structure, operator interfaces, and causal 
and resource constraints, while allowing the concrete strategies within these interfaces to be 
explored and extended during discovery. We first formalize the query-agnostic streaming task, 
then define the memory-program language and its execution semantics, and finally describe how 
an LLM proposes and refines programs through experimental feedback.

\subsection{Query-Agnostic Streaming Task}

Let $\mathcal V=\{x_1,x_2,\ldots\}$ denote an open-ended video stream and let query
$q_\tau$ arrive at an unknown future time $\tau$. Before the query is revealed, 
a memory program $p$ must causally update its complete online state:
\begin{equation}
S_t^p=\mathcal U_p(S_{t-1}^p,x_t),\qquad
S_t^p=(M_t^p,Z_t^p),\quad t\leq\tau,
\end{equation}
where $M_t^p$ is the reader-facing memory and $Z_t^p$ contains bounded working state, 
such as pending segments, previous observations, \etc 
The update processes each frame once and cannot revisit the original video. Importantly, 
its interface exposes neither future observations nor the query.
At a query cutoff, the current state is cloned and finalized into an end-of-prefix snapshot:
\begin{equation}
\bar M_\tau^p=\mathcal F_p(S_\tau^p),
\end{equation}
where $\mathcal F_p$ may flush bounded pending state. Once $q_\tau$ is revealed, 
query-conditioned retrieval and packing are allowed:
\begin{equation}
C_\tau^p=
\mathcal P_p(\bar M_\tau^p,q_\tau;B_{\mathrm{ctx}}),
\qquad
\hat y_\tau=F_\theta(q_\tau,C_\tau^p),
\end{equation}
where $B_{\mathrm{ctx}}$ is the reader-context budget, $C_\tau^p$ is the packed context, 
and $F_\theta$ is the frozen MLLM. The optimization target is therefore the program $p$ 
that maintains and reads memory, rather than the memory contents produced for any 
individual video.

\subsection{A DSL for Memory Mechanisms}

\paragraph{Program structure.}
The DSL represents a memory mechanism as a small stateful program whose reader-facing state is
$M_t^p=(R_t,E_t^p,G_t^p)$.
$R_t$ is a fixed recent window containing the latest $H$ frames, $E_t^p$ contains sparse 
high-fidelity episodic evidence, and $G_t^p$ contains compressed gist entries representing 
longer temporal intervals. Each $E$ or $G$ entry contains a reader payload, a retrieval key, 
and temporal metadata. We include $R$ in the program representation for completeness, 
but keep its implementation and capacity fixed throughout the reported search.
\begin{equation}
\begin{aligned}
\mathbf c_{EG}(M_t^p)
=[c_E(E_t^p),c_G^{\mathrm{text}}(G_t^p),
   c_G^{\mathrm{vis}}(G_t^p)]
\preceq\mathbf B_{\mathrm{mem}}(p),\qquad \forall t.
\end{aligned}
\label{eq:typed_budget}
\end{equation}
The capacity vector $\mathbf B_{\mathrm{mem}}(p)$ is part of the program and may itself 
be modified during search. Once a candidate program is instantiated, however, its 
capacities remain fixed as the stream grows.

At the level of program structure, the DSL is
\begin{equation}
\begin{aligned}
p &::= \operatorname{Memory}(R,E,G,\operatorname{Read}),\\
R &::= \operatorname{Recent}(H),\\
E &::= \operatorname{Admit}(u_E,a)\triangleright
        \operatorname{Retain}(r,B_E),\\
G &::= \operatorname{Segment}(s)\triangleright
        \operatorname{Encode}(z)\triangleright
        \operatorname{Merge}(c,\mathbf B_G),\\
\operatorname{Read}(q) &::= \operatorname{Pack}\!\left(R,
        \operatorname{Retrieve}(E,q;k_E,\delta_E),
        \operatorname{Retrieve}(G,q;k_G,\delta_G)\right).
\end{aligned}
\label{eq:policy_space}
\end{equation}
where $\triangleright$ denotes sequential composition. Here $u_E$ determines the 
episodic payload to admit; $a$, $r$, $s$, $z$, and $c$ denote the admission, retention, 
segmentation, encoding, and consolidation strategies, respectively. The $R/E/G$ roles, 
their interfaces, the frozen models, and the context-packing order are fixed by the 
runtime. Search operates only through the exposed strategy slots and their parameters.

\paragraph{Extensible operator vocabulary.}
Crucially, Equation~\ref{eq:policy_space} specifies a fixed \emph{program skeleton and interface contracts}, 
rather than a closed enumeration of all possible operators. Let $\mathcal O_r$ denote
the operator library available after search round $r$, and let $\mathcal L(\mathcal O_r)$ 
denote the programs expressible using operators satisfying these interfaces. Search begins 
from a deliberately small library $\mathcal O_0$. The research agent may either 
compose and reparameterize existing operators or propose new strategies. A newly proposed operator is 
admitted only after it satisfies the corresponding type, access, and resource contracts, 
after which it becomes available to subsequent rounds. This distinction is important: 
the DSL constrains \emph{where} and \emph{how} a memory mechanism may change, but the 
concrete set of algorithms considered during search need not be exhaustively enumerated in advance.

\begin{figure}[H]
\centering
\begin{minipage}{0.9\linewidth}
\begin{lstlisting}[style=memorydsl]
(*@\textcolor{controlcolor}{\textbf{memory}}@*) {
  (*@\textcolor{recentcolor}{\textbf{R}}@*) := (*@\textcolor{recentcolor}{\textbf{\textit{recent}}}@*)(4)
  (*@\textcolor{eventcolor}{\textbf{E}}@*) := (*@\textcolor{eventcolor}{\textbf{\textit{admit}}}@*)(frame, hist_change(q50) | gap(16s))
       |> (*@\textcolor{eventcolor}{\textbf{\textit{retain}}}@*)(coverage_change(flow,w=0.75), cap=64)
  (*@\textcolor{globalcolor}{\textbf{G}}@*) := (*@\textcolor{globalcolor}{\textbf{\textit{segment}}}@*)(fixed(8))
       |> (*@\textcolor{globalcolor}{\textbf{\textit{encode}}}@*)(text(128) + pooled_visual(1/4,128))
       |> (*@\textcolor{globalcolor}{\textbf{\textit{merge}}}@*)(oldest_adjacent, caps=(512 text,2048 visual))

  (*@\textcolor{controlcolor}{\textbf{read}}@*)(q) := (*@\textcolor{controlcolor}{\textbf{\textit{pack}}}@*)((*@\textcolor{recentcolor}{\textbf{R}}@*), (*@\textcolor{eventcolor}{topk(\textbf{E},q,8)}@*), (*@\textcolor{globalcolor}{topk(\textbf{G},q,4)}@*))
}
\end{lstlisting}
\end{minipage}
\caption{DSL representation of the selected memory program.
The pipe composes operations; \texttt{q50} is the median histogram-change threshold.
Encoding caps are per entry, while merge caps are global.}
\label{fig:memory_program}
\end{figure}

\paragraph{Execution semantics and validity.}
On each incoming frame, $R$ retains the latest $H$ observations. When the episodic 
admission condition is satisfied, an $E$ entry is constructed from $u_E$; $\operatorname{Retain}$ 
then enforces the declared capacity. For gist memory, $\operatorname{Segment}$ maintains a bounded 
pending segment. Once it closes, a frozen writer produces capped text and, 
when specified by $z$, pooled visual tokens. If the budget is exceeded, 
$\operatorname{Merge}$ selects entries to consolidate, updates their temporal interval
and recomputes corresponding retrieval representation.
At read time, $\operatorname{Retrieve}$ independently ranks $E$ and $G$ entries by cosine similarity 
between query and keys, optionally applies thresholds $\delta_E$ and $\delta_G$, 
and returns at most $k_E$ and $k_G$ entries. $\operatorname{Pack}$ restores temporal 
order within each lane and packs entries in the fixed order $R\rightarrow E\rightarrow G$ 
under $B_{\mathrm{ctx}}$. 
Preflight checks syntax, type compatibility, and declared resources of candidate programs 
before evaluation, while the runtime enforces access and capacity constraints during 
execution. These restrictions yield a structured space of executable memory mechanisms 
rather than unrestricted source-code generation.
Figure~\ref{fig:memory_program} shows the program ultimately selected by the search.

\begin{algorithm}[H]
\small
\caption{Plan-driven search over memory programs}
\label{alg:policy_discovery}
\begin{algorithmic}[1]
\Require DSL grammar $\mathcal G$, operator library $\mathcal O_0$, evaluation pipeline,
validation set $\mathcal D$, stage budgets $\{B_s\}$
\Ensure Selected program $p^\star$
\State Evaluate reference program (or import verified results); initialize history $\mathcal H$
\State $\mathcal O\gets\mathcal O_0$
\For{$s\in[E,\mathrm{Retrieval},G]$}
    \State Fix non-target program components; initialize remaining budget $B_{s,\mathrm{rem}}$
    \While{$B_{s,\mathrm{rem}}>0$ and a refinement round is proposed}
        \State $\mathcal A_s\gets\operatorname{Pareto}(\text{successful stage-compatible trials in }\mathcal H)$
        \State $\mathcal T\gets\operatorname{Plan}_{\mathrm{LLM}}(\mathcal L_s,\mathcal H,\mathcal A_s,B_s)$
        \State Freeze $\mathcal T$; stop if it contains no new candidates
        \State $\Delta\mathcal O\gets\operatorname{Implement}{\mathrm{LLM}}(\text{new operators required by }\mathcal T)$
        \State $\Delta\mathcal O\gets\operatorname{PreflightOperators}(\Delta\mathcal O)$; $\mathcal O\gets\mathcal O\cup\Delta\mathcal O$
        \State $\mathcal C\gets\operatorname{PreflightPrograms}(\operatorname{Deduplicate}(\operatorname{Compile}(\mathcal T,\mathcal O)))$
        \ForAll{$p\in\mathcal C$ \textbf{in parallel}}
            \State $(\mathbf m_p,\ell_p)\gets\operatorname{Eval}(p;\mathcal D,F_\theta)$
            \State Record $(p,\mathbf m_p,\ell_p)$ in $\mathcal H$
        \EndFor
        \State Rank successful candidates; update anchors and $B_{s,\mathrm{rem}}$
    \EndWhile
    \State Freeze the selected stage anchor
\EndFor
\State Freeze top-ranked program as $p^\star$ for external evaluation
\State \Return $p^\star$
\end{algorithmic}
\end{algorithm}

\paragraph{Program evaluation.}
Each candidate is evaluated with the same frozen models and evaluation harness. 
On the development set, let $Q(p)$ denote the mean probability assigned to the gold answer. We additionally measure
payload storage, packed reader-context length and reader latency:

\begin{equation}
    \mathbf f(p)=\big(
    Q(p),-C_{\mathrm{store}}(p),
    -C_{\mathrm{ctx}}(p),-C_{\mathrm{lat}}(p)
    \big).
    \label{eq:pareto}
\end{equation}

Successful candidates are summarized by an empirical Pareto archive over these measured 
objectives. The archive is used to expose quality--cost trade-offs within the programs 
actually evaluated; it is not intended to approximate the global Pareto frontier of 
the full program space. Failed runs are retained separately in the search history as feasibility feedback.

\subsection{LLM-Guided Program Search}

The DSL turns memory design into a structured program-search problem. Its choices 
are not merely interchangeable scalar hyperparameters, and the operator vocabulary
can be extended during search process.
We therefore use an LLM as a semantics-aware proposal model. At each round, the agent 
receives the current DSL interfaces, available operators, accumulated experimental 
history, empirical Pareto archive, and remaining evaluation budget. It first proposes 
a finite search plan $\mathcal T_r$ describing candidate program edits and the 
comparisons they are intended to test.
If a proposed candidate uses only existing operators, it can be compiled directly. 
If the plan introduces a new strategy, the LLM implements this corresponding DSL 
operator. The evaluation harness itself remains fixed throughout 
search. Candidate programs and newly introduced operators are deduplicated 
and preflighted before being executed. Their measured quality, cost, and failure 
diagnostics are returned to the agent and become context for next rounds. Thus, 
the LLM proposes what to try, while experimental execution determines whether the proposal is useful.
Algorithm~\ref{alg:policy_discovery} summarizes this process.

Rather than enumerating the full product of program choices, we use a staged
search schedule $E\rightarrow\mathrm{Retrieval}\rightarrow G$. It
is fixed by the discovery procedure rather than being searched. 
The $E$ stage evaluates the target component with the $R{+}E$ reader configuration;
retrieval and $G$ stages use $R{+}E{+}G$, with previously selected components held 
fixed. Within each stage, search proceeds from broader comparisons to local refinements 
around up to three promising anchors. Failed trials remain in the history as 
information about infeasible or ineffective designs, while only successful candidates 
enter the empirical Pareto archive.
After discovery, deployment executes only the selected memory program, 
its query-time readout, and the frozen MLLM reader; no LLM search overhead is 
introduced into the serving-time stream.

\begin{figure}[H]
    \centering
    \includegraphics[width=0.99\textwidth]{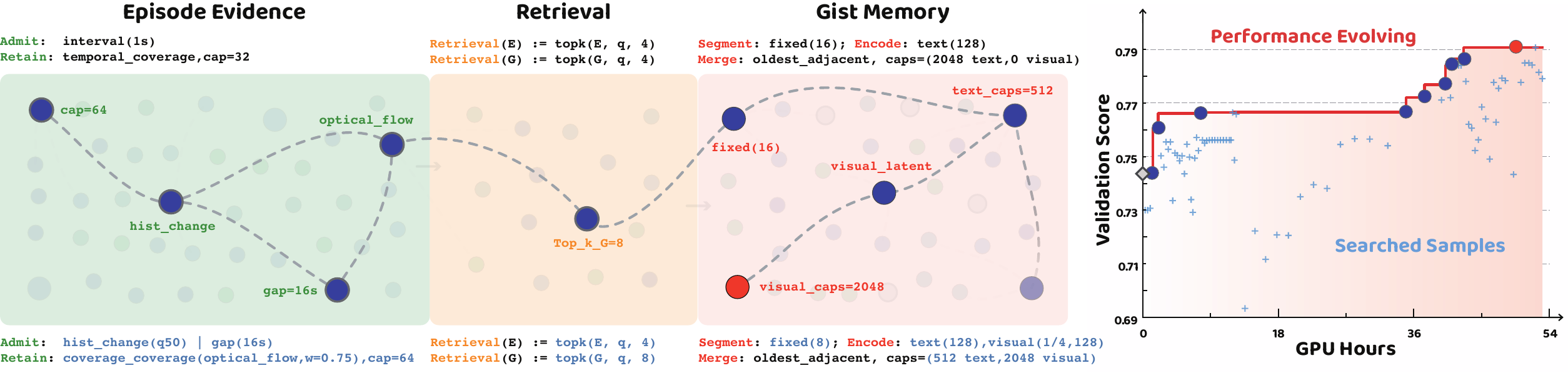}
    \caption{Automated research process of \sys. 
    The left panel visualizes the three stages of the search process. 
    Each node represents a program explored during the search, with darker 
    colors indicating better performance on validation set. 
    The dash line connects key milestones, while detailed search paths are hidden.
    At the end of each stage, the corresponding component of the 
    memory program evolves from the initial program shown at the top to the 
    final one shown at the bottom. The major milestones are also marked 
    at the corresponding positions in the evolution plot on the right.
    }
    \label{fig:search}
\end{figure}

\section{Experiments}
\label{sec:exp}
\subsection{Experimental Setup}

\paragraph{Program Discovery.}
We use NExT-GQA~\citep{next-gqa} as the development benchmark for memory-program search. 
Its questions frequently require evidence from different moments of a video, making performance
sensitive to what historical information is preserved and retrieved. We use its validation split,
containing 3,358 multiple-choice questions over 567 videos, and optimize the mean probability 
assigned to the gold option. Qwen3-VL-8B~\citep{qwen3-vl} is used for memory construction and 
as the frozen reader, while Qwen3-VL-Embedding-2B provides retrieval embeddings. Recent memory 
is fixed to 4 frames. The research agent is powered by GPT-5.6~\citep{gpt5.6}, and the complete 
discovery process takes approximately 54 GPU-hours on an NVIDIA A100 80GB GPU. The selected program 
is frozen before evaluation on the external streaming benchmarks.

\paragraph{Evaluation Benchmark.}
We evaluate on StreamingBench~\citep{StreamingBench} and OVO-Bench~\citep{OVO-Bench}. 
StreamingBench contains 900 videos and 4,500 human-curated question-answer pairs with 
queries issued at different timestamps; we use its Real-Time Visual Understanding (RTVU) 
subset. For OVO-Bench, we focus on Real-Time Visual Perception (RT), which evaluates 
current-scene perception under accumulated history, and Backward Tracing (BT), which 
evaluates retention and retrieval of information from the distant past.

\paragraph{Baselines.}
We evaluate \sys with Qwen2.5-VL-7B and Qwen3-VL-8B backbones at 1 FPS. Training-based 
baselines include VideoLLM-Online~\citep{VideoLLM-online}, Flash-VStream~\citep{Flash-VStream}, 
Dispider~\citep{Dispider}, TimeChat-Online~\citep{TimeChat-Online}, and StreamForest~\citep{StreamForest}. 
Training-free baselines include StreamBridge~\citep{StreamBridge}, InfiniPot-V~\citep{InfiniPot-V}, 
HERMES~\citep{HERMES}, FluxMem~\citep{FluxMem}, and OASIS~\citep{OASIS}. Detailed backbone and 
result sources are provided in Appendix~\ref{app:baseline-settings}.

\subsection{Memory Program Discovery}
\label{sec:discovery_analysis}

\paragraph{LLM-guided search progressively improves the memory program.}
Figure~\ref{fig:search} summarizes the discovery trajectory. Starting from a
simple reference program, \sys searches episodic evidence, query-time retrieval, 
and gist memory sequentially, freezing the selected component before moving to the 
next stage. Rather than enumerating the Cartesian product of possible designs, each 
round uses previous quality, efficiency, and failure measurements to determine which 
program edits should be evaluated next. 
As shown by the staircase plot on the right, the research agent can accurately identify
the key milestones and outcomes at each stage, continually pushing the performance frontier.
Over approximately 54 GPU-hours, the discovered
program improves the development objective by 5--6 points.
The selected program is depicted as Figure~\ref{fig:memory_program}.

\begin{table}[H]
    \centering\small
    \caption{Experiment results on subsets of StreamingBench and OVO-Bench. The highest score
        in each column is marked with bold, and the second highest with underline.}
    \label{tab:streaming_results}
    \renewcommand{\arraystretch}{1.1}
    \begin{tabular}{lcccccc}
        \toprule
         &                                                            &  & \multicolumn{1}{c}{\textbf{StreamingBench}}
         & \multicolumn{3}{c}{\textbf{OVO-Bench}}                                                                      \\
        \cmidrule(lr){4-4}
        \cmidrule(lr){5-7}
        \textbf{Method}
         & \textbf{Size}
         & \textbf{\#Frames}
         & \textbf{RTVU}
         & \textbf{RT}
         & \textbf{BT}
         & \textbf{Avg.}                                                                                               \\
        \midrule

        \rowcolor{sectionbg}
        \multicolumn{7}{c}{\textit{\textbf{Training-based Methods}}}                                                   \\
        \midrule

        VideoLLM-online
         & 8B
         & 2 fps
         & 36.0
         & 20.8
         & 17.7
         & 19.3                                                                                                        \\

        Flash-VStream
         & 7B
         & 1 fps
         & 23.2
         & 28.4
         & 27.4
         & 27.9                                                                                                        \\

        Dispider
         & 7B
         & 1 fps
         & 67.6
         & 54.6
         & 36.1
         & 45.4                                                                                                        \\

        TimeChat-Online
         & 7B
         & 1 fps
         & 75.4
         & 61.9
         & 41.7
         & 51.8                                                                                                        \\

        StreamForest
         & 7B
         & 1 fps
         & 77.3
         & 61.2
         & 52.0
         & 56.6                                                                                                        \\

        \midrule
        \rowcolor{sectionbg}
        \multicolumn{7}{c}{\textit{\textbf{Training-free Methods}}}                                                    \\
        \midrule


        StreamBridge
         & 7B
         & 1 fps
         & 72.0
         & 63.4
         & \textbf{59.5}
         & 61.5                                                                                                        \\

        InfiniPot-V
         & 7B
         & 1 fps
         & 76.4
         & 65.9
         & 47.6
         & 56.8                                                                                                        \\

        FluxMem
         & 7B
         & 1 fps
         & 76.4
         & 67.2
         & 47.2
         & 57.2                                                                                                        \\

        HERMES~{\tiny(qwen2.5vl)}
         & 7B
         & 1 fps
         & 79.4
         & 69.0
         & 49.4
         & 59.2                                                                                                        \\

        HERMES~{\tiny(qwen3vl)}
         & 8B
         & 2 fps
         & \underline{81.3}
         & 73.3
         & 49.3
         & 61.3                                                                                                        \\

        OASIS~{\tiny(qwen2.5vl)}
         & 7B
         & 0.5 fps
         & 70.6
         & 67.3
         & 52.6
         & 60.0                                                                                                        \\

        OASIS~{\tiny(qwen3vl)}
         & 8B
         & 0.5 fps
         & 78.2
         & \underline{78.1}
         & 57.2
         & 67.7                                                                                                        \\

        Qwen2.5-VL
         & 7B
         & 1 fps
         & 73.3
         & 59.9
         & 44.7
         & 52.3                                                                                                        \\

        \rowcolor{highlightbg}
        \textbf{\;\;+ \sys}
         & \textbf{7B}
         & {1 fps}
         & 77.5 \textbf{\tiny{\textcolor{teal}{(+4.2)}}}
         & 73.8 \textbf{\tiny{\textcolor{teal}{(+13.9)}}}
         & 52.0 \textbf{\tiny{\textcolor{teal}{(+7.3)}}}
         & 62.9 \textbf{\tiny{\textcolor{teal}{(+10.6)}}}                                                              \\

        Qwen3-VL
         & 8B
         & 1 fps
         & 75.7
         & 68.2
         & 46.4
         & 57.3                                                                                                        \\

        \rowcolor{highlightbg}
        \textbf{\;\;+ \sys}
         & \textbf{8B}
         & {1 fps}
         & \textbf{81.4 \tiny{\textcolor{teal}{(+5.7)}}}
         & \textbf{78.2 \tiny{\textcolor{teal}{(+10.0)}}}
         & \underline{59.3} \textbf{\tiny{\textcolor{teal}{(+12.9)}}}
         & \textbf{68.8} \textbf{\tiny{\textcolor{teal}{(+11.5)}}}                                                     \\

        \bottomrule
    \end{tabular}
\end{table}

\paragraph{The search discovers non-trivial interactions between memory operations.}
The reference program stores one episodic frame per second with a capacity of 32 and 
represents gist memory using 16-frame textual segments under a 2,048-token budget. 
During discovery, the agent explores alternative admission, capacity, retrieval, segmentation, 
representation, and consolidation strategies. The resulting trajectory is not simply driven 
by increasing memory size: the search favors deeper retrieval from episodic evidence than 
from gist memory, reduces the gist text budget, shortens its temporal segmentation, and 
eventually introduces pooled visual representations. 
When exploring episodic evidence memory, the research agent retains two additional settings
that are not optimal in scores but exhibit better inference efficiency,
and launches an extra exploration process to further investigate their potential. When searching
the gist impression memory, the research agent initiates multiple parallel and independent
exploration trajectories from the same starting point, thereby actively expanding the search space.
These outcomes show that representation, 
storage, and readout choices interact, and that more memory or more complex operators do not 
necessarily yield better downstream performance. Detailed program changes and intermediate 
measurements are provided in Appendix~\ref{app:research-details}.

\subsection{Main Results on Streaming Video Benchmarks}

\paragraph{\sys consistently improves frozen MLLM backbones and achieves superior performance.}
Table~\ref{tab:streaming_results} summarizes the main results. With Qwen2.5-VL-7B, \sys improves RTVU by 4.2 
points and the OVO-Bench average by 10.6 points. With Qwen3-VL-8B, it reaches 81.4 on RTVU 
and 68.8 on OVO-Bench, corresponding to gains of 5.7 and 11.5 points over the frozen backbone, 
respectively. These are the highest overall scores among the evaluated methods. On OVO-Bench, 
\sys reaches 78.2 on RT and 59.3 on BT, indicating that the discovered memory program improves 
both immediate perception and access to historical information.

A detailed fine-grained analysis on StreamingBench is performed with results in Table~\ref{tab:streamingbench_finegrained_results}.
It further shows that the aggregate gain is distributed across a broad range of capabilities. 
With Qwen3-VL-8B, \sys ranks first in 7 out of 10 subcategories and improves over its backbone in eight. The largest
gains occur on Clips Summarization and Text-Rich Understanding, where the scores increase from
78.6 to 92.7 and from 80.3 to 91.1, corresponding to improvements of 14.1 and 10.8 points.
Although Causal Reasoning and Counting remain below the unaugmented backbone, the breadth of
the remaining gains indicates that the overall improvement is not driven by a single subcategory.

\begin{table}[H]
    \centering\small
    \caption{Fine-grained subcategory results on StreamingBench Real-Time subset.
        Baseline scores are officially reported values. The abbr for subcategories stand for:
        \textit{Object Perception}, \textit{Causal Reasoning}, \textit{Clips Summarization}, \textit{Attribute Perception},
        \textit{Event Understanding}, \textit{Text-Rich Understanding}, \textit{Prospective Reasoning}, \textit{Spatial Understanding},
        \textit{Action Perception} and \textit{Counting}.}
    \label{tab:streamingbench_finegrained_results}
    \renewcommand{\arraystretch}{1.1}
    \begin{tabular}{lccccccccccc}
        \toprule
        \textbf{Method}
         & \textbf{OP}
         & \textbf{CR}
         & \textbf{CS}
         & \textbf{ATP}
         & \textbf{EU}
         & \textbf{TR}
         & \textbf{PR}
         & \textbf{SU}
         & \textbf{ACP}
         & \textbf{CT}
         & \textbf{Avg.}                                              \\
        \midrule

        \rowcolor{sectionbg}
        \multicolumn{12}{c}{\textit{\textbf{Training-based Methods}}} \\
        \midrule

        VideoLLM-Online
         & 39.1
         & 40.1
         & 34.5
         & 31.1
         & 46.0
         & 32.4
         & 31.5
         & 34.2
         & 42.5
         & 27.9
         & 36.0                                                       \\

        Flash-VStream
         & 25.9
         & 43.6
         & 24.9
         & 23.9
         & 27.3
         & 13.1
         & 18.5
         & 25.2
         & 23.9
         & 48.7
         & 23.2                                                       \\

        Dispider
         & 74.9
         & 75.5
         & 74.1
         & 73.1
         & 74.4
         & 59.9
         & 76.1
         & 62.9
         & 62.2
         & 45.8
         & 67.6                                                       \\

        TimeChat-Online
         & 80.8
         & 79.7
         & 80.8
         & 83.3
         & 74.8
         & 78.8
         & 78.7
         & 64.2
         & 68.8
         & \textbf{58.0}
         & 75.3                                                       \\

        StreamForest
         & 83.1
         & \textbf{82.8}
         & 82.7
         & 84.3
         & 77.5
         & 78.2
         & 76.9
         & 69.1
         & \underline{75.6}
         & \underline{54.4}
         & 77.3                                                       \\

        \midrule
        \rowcolor{sectionbg}
        \multicolumn{12}{c}{\textit{\textbf{Training-free Methods}}}  \\
        \midrule

        StreamBridge
         & 80.4
         & 78.7
         & 83.2
         & 79.9
         & 74.2
         & 69.5
         & 77.8
         & 63.4
         & 70.0
         & 43.0
         & 72.0                                                       \\

        FluxMem
         & 80.2
         & 81.1
         & 81.4
         & 85.3
         & \underline{78.0}
         & 83.8
         & 80.6
         & 65.9
         & 69.6
         & 52.1
         & 76.4                                                       \\

        HERMES
         & 83.7
         & \underline{81.3}
         & 88.0
         & \underline{87.5}
         & 76.7
         & 86.6
         & 82.4
         & \textbf{76.0}
         & 73.9
         & 46.6
         & \underline{79.4}                                           \\

        Qwen2.5-VL
         & 77.9
         & 76.6
         & 78.6
         & 80.9
         & 76.7
         & 77.0
         & 80.6
         & 65.5
         & 65.7
         & 52.9
         & 73.3                                                       \\

        \rowcolor{highlightbg}
        \textbf{\;\;+ \sys}
         & \underline{84.8}
         & 70.1
         & \underline{88.3}
         & 86.8
         & 75.9
         & \underline{88.5}
         & 79.6
         & 72.8
         & 67.5
         & 41.0
         & 77.5                                                       \\

        Qwen3-VL
         & 78.3
         & 75.0
         & 78.6
         & 84.6
         & 70.4
         & 80.3
         & \underline{83.3}
         & 71.5
         & 70.5
         & 51.1
         & 75.7                                                       \\

        \rowcolor{highlightbg}
        \textbf{\;\;+ \sys}
         & \textbf{86.9}
         & 69.8
         & \textbf{92.7}
         & \textbf{89.1}
         & \textbf{79.7}
         & \textbf{91.1}
         & \textbf{88.0}
         & \underline{75.6}
         & \textbf{76.2}
         & 45.5
         & \textbf{81.4}                                              \\

        \bottomrule
    \end{tabular}
\end{table}

\begin{figure}[H]
    \centering
    \begin{subfigure}[b]{0.46\textwidth}
        \centering
        \includegraphics[width=\linewidth]{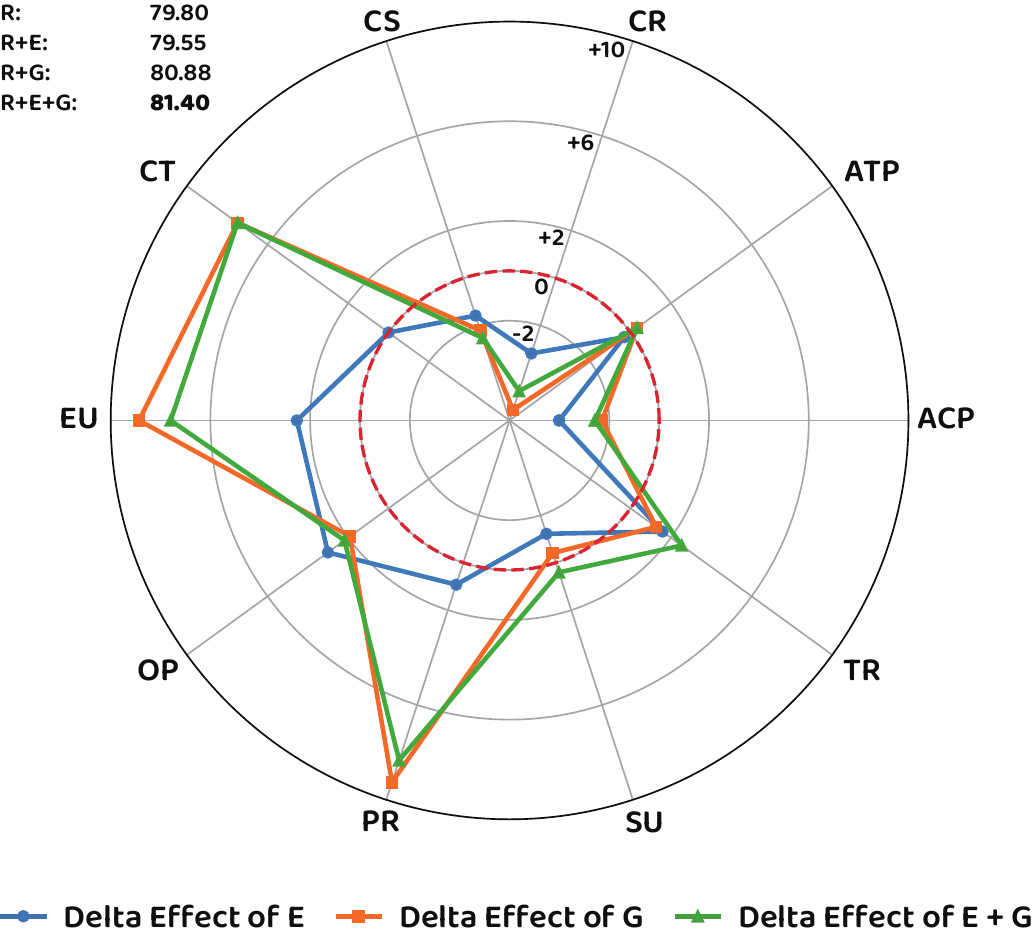}
        \caption{Marginal effects of different roles.}
        \label{subfig:lane-ablation}
    \end{subfigure}
    \begin{subfigure}[b]{0.46\textwidth}
        \centering
        \includegraphics[width=\linewidth]{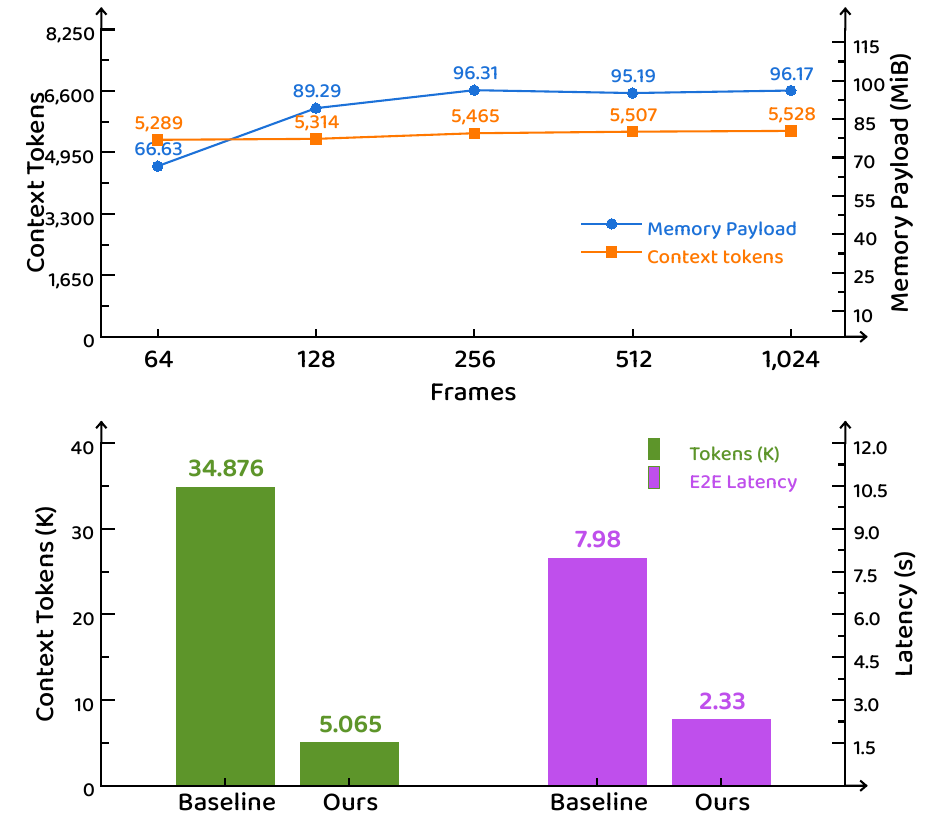}
        \caption{Computational resource profiling comparison.}
        \label{subfig:bounded-memory}
    \end{subfigure}
    \caption{Left figure denotes the marginal effects of \sys's different roles on StreamingBench subsets. 
        The red dashed circle denotes the baseline of only adding
        recent memory. The right upper denotes the trends of context tokens and memory payload
        as the frames accumulate; and the right lower denotes average token cost and latency
        on OVO-Bench compared with uniform sampling.}
    \label{fig:ablations}
\end{figure}

\paragraph{Different memory roles provide complementary information.}
Figure~\figsubref{fig:ablations}{subfig:lane-ablation} decomposes the contribution of recent ($R$), episodic ($E$), 
and gist ($G$) memory on StreamingBench. 
The red dashed circle denotes the performance baseline when only recent memory is retained.
The other colors, as indicated in the legend, show the relative gains achieved by augmenting
it with $E$, $G$, or combined.
The figure clearly shows that different memory roles contribute differently across task types.
Episodic evidence provides clear gains on Object Perception, 
Event Understanding, and Prospective Reasoning, consistent with its role in preserving sparse 
high-fidelity observations. Gist memory contributes more strongly to tasks such as Counting, 
Event Understanding, and Prospective Reasoning, which depend on information accumulated over longer temporal ranges.
The ablation also shows that adding memory is not always beneficial. On several categories (CR/ACP), 
additional historical evidence can degrade performance. Besides, we find that combining episodic and gist memory 
can recover part of this loss, reaching the highest score. This supports the central motivation for program search: the 
effectiveness of a memory mechanism depends not only on how much information is retained, 
but also on how different forms of memory are represented and combined.
Detailed case study is at Appendix~\ref{app:case-study}.

\paragraph{\sys maintains strong efficiency with modest resource overhead and bounded cost.}
We profile the discovered Qwen3-VL program (reader-side) on a single A100 GPU, as shown in 
Figure~\figsubref{fig:ablations}{subfig:bounded-memory}. 
First, to verify that the memory produced
by \sys remains stable and bounded as the video stream grows, for StreamingBench 
videos longer than 1,024 frames, we construct memory from prefixes of 64/128/256/512/1024
frames. The token counts and the locally stored payload size are recorded. 
Both reader-context size and locally stored payload gradually stabilize (5500 tokens, 96 MiB) as the 
observed stream grows, consistent with the bounded capacities enforced by the memory program.
On OVO-Bench, we further compare against directly processing uniformly sampled 
video history with the frozen backbone. \sys reduces the average reader context 
from nearly 35K tokens to approximately 5K tokens, an 86\% reduction, while 
reducing end-to-end inference latency by 71\%. More profiling details and results are provided in Appendix~\ref{app:profiling}.
This demonstrates the \sys memory structure enables efficient inference under a fixed memory budget.

\section{Conclusion}

We presented \sys, an LLM-driven framework for automatically discovering 
memory mechanisms for query-agnostic streaming video understanding. Rather 
than manually designing a single memory architecture, \sys represents memory 
mechanisms as executable programs in a compact DSL with explicit causal and 
bounded-resource constraints. An LLM acts as a semantics-aware proposal model 
that iteratively explores program choices and uses feedback from fixed runtime
to refine the design. The reseach agent will not participate in the external
benchmark evaluation after the search ends and optimal program is out. Experiments on StreamingBench and
OVO-Bench show that the discovered program substantially improves frozen backbones 
while maintaining bounded memory and efficient inference. Further analysis demonstrates 
complementary roles of recent, episodic, and gist memory and reveals that effective 
memory design depends on interactions among representation, storage, and readout 
choices. These results suggest that automated memory-program discovery is a 
practical alternative to manually engineering a single streaming memory mechanism.


\section*{AI Use Statement}

We used generative AI tools to assist with both manuscript preparation
and research ideation and execution. For manuscript preparation, these
tools supported drafting and revising portions of the text to improve
clarity, organization, and language. For research, they assisted with
experimental implementation, debugging, and interpretation of results.
In particular, the GPT-5.6 research agent in \sys proposed and refined
candidate memory programs and implemented new DSL operators using
experimental feedback, as described in Section~\ref{sec:method} and
Appendix~\ref{app:research-details}. Candidate programs were evaluated
using a fixed evaluation harness, with measured performance, resource
costs, and execution failures informing subsequent search.

The authors reviewed and edited AI-assisted text, inspected AI-assisted
code, and checked analyses against experimental outputs. Research
proposals were assessed through controlled experiments. We did not use
generative AI tools to create synthetic datasets or prove mathematical
claims. The authors take full responsibility for the final manuscript,
including its claims, results, code, and other AI-assisted artifacts.

\bibliography{refs}
\bibliographystyle{iclr2027_conference}

\newpage
\appendix
\section{Detailed Evolution Process of Automated Research}
\label{app:research-details}

This section provides the complete evolution process of the automated
memory-program search. As described in Section~\ref{sec:method}, \sys does not
optimize the stored contents of individual videos. Instead, it searches
over executable programs in the memory DSL by modifying the exposed operators
and parameters while preserving the fixed program skeleton, causal interfaces,
and resource constraints.

Figure~\ref{fig:search} visualizes all candidates evaluated during the search.
Starting from an imported reference program, the research agent conducted 13
refinement rounds across three stages, following the fixed schedule \(E \rightarrow \mathrm{Retrieval} \rightarrow G\).
In total, 89 canonically distinct candidate programs were evaluated:
40 for episodic memory, 15 for query-time retrieval, and 34 for gist memory.
Among them, 85 completed successfully and four failed. Canonically equivalent
programs were deduplicated before evaluation.
In particular, the $E$ stage is
selected using its designated $R{+}E$ evaluation arm, whereas the Retrieval and
$G$ stages use $R{+}E{+}G$.

Table~\ref{tab:search-details} summarizes the 13 rounds in terms of the DSL component being
modified. The important feature of this trajectory is that the search is not
a greedy traversal of a fixed hyperparameter grid. Different rounds expose
different program semantics---for example, admission, retention, segmentation,
representation, or retrieval---and later rounds can recombine useful decisions
discovered in different branches.

\begin{table}[H]
    \centering
    \scriptsize
    \setlength{\tabcolsep}{4.5pt}
    \begin{tabular}{llccL{20em}}
        \toprule
        Stage & Search round / DSL edit                                             & Trials   & Best Score & Main outcome \\
        \midrule
        $E$
              & E1: $\operatorname{Retain}(r,B_E)$: capacity $\times$ retention
              & 11                                                                  & 0.748356
              & Explore capacity and FIFO/temporal/reservoir retention                                                             \\

              & E2: $\operatorname{Admit}(a)$
              & 12                                                                  & 0.752949
              & Compare interval, histogram, SSIM, and optical-flow triggers                                                       \\

              & E3: admission gap constraints
              & 5                                                                   & 0.753435
              & Refine histogram-$q50$ admission with minimum/maximum gaps                                                         \\

              & E4: $\operatorname{Retain}(r,B_E)$: coverage-change retention
              & 9                                                                   & 0.753435
              & Compare histogram/SSIM/flow coverage criteria and weights                                                          \\

              & E5: local program recombination
              & 3                                                                   & 0.740043
              & Recombine promising admission, gap, retention, and capacity choices                                                \\
        \midrule
        Retrieval
              & $\operatorname{Retrieve}(E,q;k_E)$,
        $\operatorname{Retrieve}(G,q;k_G)$
              & 15                                                                  & 0.771990
              & Select $(k_E,k_G)=(8,4)$                                                                                           \\
        \midrule
        $G$
              & G1: $\operatorname{Segment}(s)$ + text $\operatorname{Encode}(z)$
              & 8                                                                   & 0.784083
              & Favor short fixed segments; three low-text-budget trials fail                                                      \\

              & G2: global text budget $\mathbf B_G$
              & 3                                                                   & 0.786595
              & Reduce global text cap to 512                                                                                      \\

              & G3: $\operatorname{Merge}(c)$
              & 2                                                                   & 0.778048
              & Alternative consolidation does not improve the current anchor                                                      \\

              & G4: boundary-based $\operatorname{Segment}(s)$
              & 9                                                                   & 0.777711
              & Test histogram/SSIM/flow-based adaptive segmentation                                                               \\

              & G5: local boundary refinement
              & 4                                                                   & 0.779277
              & Refine minimum and maximum segment lengths                                                                         \\

              & G6: hybrid $\operatorname{Encode}(z)$
              & 4                                                                   & 0.790721
              & Add pooled visual tokens to the textual gist                                                                       \\

              & G7: local hybrid refinement
              & 4                                                                   & 0.790721
              & Refine visual pooling and visual-token budgets                                                                     \\
        \bottomrule
    \end{tabular}
    \caption{
        Round-by-round evolution of the automated search.}
    \label{tab:search-details}
\end{table}

\subsection{Evolution of Episodic Memory}
The first five rounds modify the episodic branch 
\[
E :=
\operatorname{Admit}(u_E,a)
\triangleright
\operatorname{Retain}(r,B_E),
\]
while leaving the other DSL components fixed. Rather than proposing a single 
episodic mechanism at once, the agent progressively explores storage capacity, 
retention behavior, admission rules, and their local interactions.

\paragraph{E1: capacity and retention.}
The first round probes the basic trade-off between episodic capacity $B_E$ 
and the retention operator $r$. Eleven programs combine capacities from 8 to 
64 entries with FIFO, temporal, and reservoir-style retention. Very small 
capacities consistently perform poorly under the metric: the three 
capacity-8 variants obtain approximately $0.686$-$0.689$. Increasing capacity improves 
performance, with \texttt{e-cap16-reservoir} reaching $0.735083$.
Importantly, subsequent exploration is not restricted to the single node 
with the highest score. The search retains multiple 
candidate anchors according to the stage-specific objective and accumulated 
quality--cost feedback. This behavior allows later rounds to investigate 
designs that may interact differently with admission and retention operators.

\paragraph{E2: admission strategies.}
The second round edits admission strategy $a$. Twelve programs compare 
fixed-interval admission against content-dependent triggers derived from 
histogram change, SSIM, and optical flow, with several quantile thresholds. 
The results reveal substantial sensitivity to the admission rule: aggressive 
high-threshold variants such as histogram-$q90$ and optical-flow-$q90$ obtain 
scores of $0.697551$ and $0.695995$, respectively, whereas 
optical-flow-$q75$ reaches $0.730817$.
The agent does not simply freeze this locally strongest trigger. Instead, 
histogram-$q50$ is retained as a promising branch for further structural 
refinement, illustrating that candidate generation operates over program 
semantics rather than greedily following a single scalar observation.

\paragraph{E3: temporal gap constraints.}
Starting from histogram-$q50$ admission, the third round augments the admission 
program with explicit temporal-gap constraints. Five candidates test minimum-gap 
variants and maximum-gap variants. A minimum gap of 2 reaches a new
running best of $0.739975$, while maximum-gap variants remain around $0.7527$--$0.7534$.
These trials expose a useful distinction between two admission behaviors: a minimum
gap suppresses temporally redundant evidence, whereas a maximum gap guarantees 
that sufficiently long periods cannot pass without an admitted observation. The 
latter can therefore provide temporal coverage even when appearance-change 
triggers remain inactive.

\paragraph{E4: coverage-aware retention.}
The fourth round moves from admission to retention. Starting from the histogram-$q50$ 
branch with a maximum gap of 16, the agent introduces coverage-change retention strategies 
based on histogram, SSIM, or optical-flow statistics, each with weights $0.25$, $0.50$, 
and $0.75$. All nine programs obtain the same score of $0.753435$ 
on the validation set, so this visualization alone does not distinguish their retention behavior.
Under the designated episodic-stage evaluation and selection procedure, the optical-flow 
coverage strategy with weight $0.75$ is frozen as the $E$-stage winner. Its resulting 
DSL branch can be summarized as
\[
\operatorname{Admit}
\bigl(
\mathrm{hist\_change}(q50),\lor,\mathrm{gap}(16\mathrm{s})
\bigr)
\triangleright
\operatorname{Retain}
\bigl(
\mathrm{coverage\_change}(\mathrm{flow},w=0.75),
B_E=64
\bigr).
\]

\paragraph{E5: local recombination.}
Before closing the episodic stage, the agent evaluates three local combinations 
intended to recombine promising decisions from previous branches. Combining the 
histogram-$q50$ admission with the minimum-gap-$2$ branch yields scores 
of $0.740043$ and $0.740008$ depending on the maximum-gap setting, while reducing 
the coverage-based program to capacity 32 obtains $0.738682$.
Finally, the selected episodic program in E4 is then held fixed in all subsequent stages.

\subsection{Evolution of Query-Time Retrieval}
This stage evaluates 15 canonically distinct programs that vary the retrieval 
depths for episodic and gist memory.
\[\operatorname{Pack}
\left(
R,
\operatorname{Retrieve}(E,q;k_E,\delta_E),
\operatorname{Retrieve}(G,q;k_G,\delta_G)
\right).
\]

The explored values span $k_E,k_G\in{1,2,4,8}$, 
with canonical duplicates removed before execution.
The resulting pattern shows that the two memory lanes benefit from different 
retrieval depths. Restricting episodic retrieval to one or two entries performs 
poorly, with scores ranging from $0.693351$ to $0.739516$. Increasing $k_E$ to 8 
produces a clear improvement: $(k_E,k_G)=(8,2)$ reaches $0.766595$, while $(8,4)$ 
reaches the stage-best score of $0.771990$. Increasing gist retrieval further to 
$k_G=8$ does not improve the result, yielding $0.771109$.
The retrieval stage therefore freezes
\[
k_E=8,\qquad k_G=4.
\]
This result also illustrates that allocating more reader context is not monotonically 
beneficial: the selected program retrieves substantially more episodic evidence 
than gist entries, and increasing the gist depth from four to eight slightly reduces the validation score.

\subsection{Evolution of Gist Memory}
With episodic memory and retrieval fixed, the final seven rounds explore
\[
G :=
\operatorname{Segment}(s)
\triangleright
\operatorname{Encode}(z)
\triangleright
\operatorname{Merge}(c,\mathbf B_G).
\]
This is the largest and most heterogeneous part of the search, covering temporal segmentation, 
textual representation, global memory capacity, consolidation, and hybrid visual--textual encoding.

\paragraph{G1: fixed segmentation and textual encoding.}
The first gist round jointly varies fixed segment length and per-entry text budget. 
Eight programs are launched. Three configurations using a text budget of 64 fail and 
are retained in the history as feasibility feedback rather than quantitative observations.
Among successful programs, shorter segments are favored. Fixed 32-frame segments with 
text budgets of 128 or 256 both obtain $0.776876$, whereas fixed 8-frame segments 
reach $0.784083$ with either 128 or 256 text tokens. The equality between the two 
latter variants suggests that increasing the per-entry textual allowance beyond 128 
provides no observable benefit on this validation set. The more compact
\texttt{fixed(8)+text(128)} is retained as one of final Pareto candidates.

\paragraph{G2: global text capacity.}
The second gist round keeps the 8-frame, 128-token representation and varies 
the global text capacity. Capacities 1024 and 4096 both preserve the previous 
score of $0.784083$, whereas reducing the global text budget to 512 improves 
the score to $0.786595$. The resulting program becomes a new running best and 
the second final gist Pareto candidate.
This is particularly informative for bounded streaming memory: allocating 
more storage does not necessarily improve downstream reasoning. In this case, 
a tighter global gist budget both restricts memory growth and yields a slightly 
better validation result.

\paragraph{G3: alternative consolidation.}
The next round challenges the current consolidation strategy. Removing consolidation
entirely produces a failed trial, while a most-similar-entry consolidation strategy 
completes successfully but decreases the score to $0.778048$. Neither
program improves the current anchor, so the search does not promote this branch.
The failed candidate is still retained in the search history. This is important 
for the automated-research loop: execution failures constrain the feasible portion 
of the DSL and become feedback for subsequent proposals, although they are excluded 
from the empirical quality--cost archive.

\paragraph{G4--G5: adaptive temporal boundaries.}
The agent next investigates whether the fixed 8-frame segmentation can be replaced 
by content-adaptive boundaries. Nine programs use histogram, SSIM, or optical-flow 
change signals at several thresholds while bounding segment lengths between 4 and 32 frames.
Among these candidates, optical-flow-$q50$ is strongest at $0.777711$, but it remains 
below the fixed-segment anchor at $0.786595$. A further local round varies its minimum
and maximum segment lengths. The best boundary refinement, with minimum 4 and maximum 64 
frames, reaches $0.779277$; reducing the minimum to one frame substantially 
degrades the score to $0.743346$.
Thus, although adaptive segmentation introduces a more elaborate DSL operator, 
the experiments provide no evidence that it improves over the simpler fixed-8 
strategy in this setting. The search consequently returns to the fixed-segment 
branch for subsequent representation exploration.

\paragraph{G6: hybrid visual--textual gist.}
The sixth gist round changes the representation operator $z$ rather than the 
segmentation rule. Starting from
\[
\operatorname{Segment}(\mathrm{fixed}(8))
\triangleright
\operatorname{Encode}(\mathrm{text}(128)),
\]
the agent augments each gist entry with pooled visual tokens and varies the 
pooling ratio, per-entry visual-token allowance, and global visual budget.
The strongest candidate uses a pooling ratio of $1/4$, a per-entry visual cap 
of 128, and a global visual budget of 1024, reaching a new running best of 
$0.790721$. More aggressive visual representations do not improve this result: 
the tested $1/2$ pooling variants obtain scores between $0.777712$ and $0.784974$. 
This indicates that a relatively compact visual payload complements the textual 
gist more effectively than simply retaining more visual tokens.

\paragraph{G7: local hybrid refinement.}
The final round performs a local search around the best hybrid representation. 
Four nearby programs vary the pooling ratio, per-entry visual cap, and global 
visual budget. Increasing the global visual cap from 1024 to 2048 while retaining 
the $1/4$ pooling ratio and 128-token per-entry visual cap preserves the best 
score of $0.790721$. Other variants that increase the per-entry visual cap to 
256 or use a denser $1/2$ representation perform worse.

According to the final stage selection, the program
\[
\operatorname{Encode}
\bigl(
\mathrm{text}(128)
+
\mathrm{pooled\_visual}(1/4,128)
\bigr),
\qquad
\mathbf B_G=(512\ \text{text},2048\ \text{visual})
\]
is frozen as the $G$-stage winner and the first final gist Pareto candidate.

\subsection{Resulting Memory Program}
The complete search trajectory progressively modifies different semantic 
slots of the DSL rather than attempting to enumerate their full Cartesian 
product. The final program combines decisions discovered across the three stages,
as shown in Figure~\ref{fig:memory_program}.

The trajectory highlights three properties of the automated search. \emph{(i)} 
useful improvements arise from \emph{coupled program edits}: admission, retention, 
representation, and retrieval cannot be optimized independently by simply increasing 
their capacities. \emph{(ii)} the search is explicitly empirical. More sophisticated 
operators, such as adaptive boundary segmentation or most-similar consolidation, 
are retained only when execution supports them; complexity itself is not treated 
as evidence of improvement. \emph{(iii)} the search is not greedily tied to a single 
intermediate score. The agent can preserve and revisit multiple semantic branches, 
recombine promising program components, and use failed executions as feasibility feedback.
After this procedure terminates, all search-time machinery is removed. The research 
agent does not participate in serving-time inference; only the frozen program, 
its bounded online state, the query-time readout, and the frozen MLLM are retained.

\section{Detailed Settings of Baselines}
\label{app:baseline-settings}
All experimental results reported in Section~\ref{sec:exp} are collected from published papers.
The base model and data source for each method are summarized in Table~\ref{tab:data_source}.

\begin{table}[H]
    \centering
    \caption{Base models, publication information, and data sources of compared methods.
    ``Official'' means the scores are from the method's own paper; otherwise the source is
    stated explicitly.}
    \label{tab:data_source}
    \begin{tabular}{lccc}
        \toprule
        \textbf{Method} & \textbf{Base Model} & \textbf{Method Source} & \textbf{Data Source} \\

        \midrule
        \multicolumn{4}{l}{\textit{Training-based Methods}}                            \\
        \midrule
        VideoLLM-Online
                        & Llama-3-8B          & CVPR 2024              & StreamForest paper   \\
        Flash-VStream
                        & Qwen2-VL-7B         & ICCV 2025              & OASIS paper          \\
        Dispider
                        & Qwen2-7B            & CVPR 2025              & OASIS paper          \\
        TimeChat-Online
                        & Qwen2.5-VL-7B       & ACM MM 2025            & HERMES paper         \\
        StreamForest
                        & Qwen2-7B            & NeurIPS 2025 Spotlight & Official             \\

        \midrule
        \multicolumn{4}{l}{\textit{Training-free Methods}}                             \\
        \midrule
        StreamBridge
                        & Qwen2-VL-7B         & NeurIPS 2025           & Official             \\
        InfiniPot-V
                        & Qwen2.5-VL-7B       & NeurIPS 2025           & Official             \\
        HERMES 4k
                        & Qwen2.5-VL-7B       & ACL 2026 Long          & Official             \\
        HERMES 4k
                        & Qwen3-VL-8B         & ACL 2026 Long          & Official             \\
        FluxMem
                        & Qwen2.5-VL-7B       & CVPR 2026              & Official             \\
        OASIS
                        & Qwen2.5-VL-7B       & CVPR 2026              & Official             \\
        OASIS
                        & Qwen3-VL-8B         & CVPR 2026              & Official             \\

        \bottomrule
    \end{tabular}
\end{table}

\section{Resource profiling of \sys}
\label{app:profiling}

\paragraph{Reader-side profiling.}Figure~\figsubref{fig:ablations}{subfig:bounded-memory} shows that the memory produced by 
\sys remains stable in size as more video frames accumulate. We also provide a fine-grained 
breakdown of reader-side latency, with the results summarized in Table~\ref{tab:detailed-profiling}.
As shown, the overall runtime remains stable for both retrieval and packing. 
Peak GPU memory usage also stays consistently around 18.9 GiB.

\paragraph{Writer-side profiling.}
We will focus on the deployment writer performance of \sys. Among the three memory roles, 
maintaining recent memory and episodic evidence involves no complex computation and only 
requires writing data to storage, allowing them to operate at well above 1 FPS. In contrast, 
maintaining gist memory requires both visual-token pooling over video frames and text 
summarization, making it the primary computational bottleneck. We therefore propose a pipelined 
processing scheme that sustains an operating rate above 1 FPS, enabling stable streaming execution.
Note that this setup is distinct from the single-A100 profiling experiment reported in Section~\ref{sec:exp}. 
\sys consists of a writer and a reader: before the query arrives, the writer continuously processes 
incoming video frames and writes them into memory; once the query arrives, the reader, implemented 
as a frozen MLLM, receives the packed context and produces the answer.

\begin{table}[H]
    \centering\small
    \caption{Fine-grained reader-side profiling result on StreamingBench.}
    \begin{tabular}{cccc}
        \toprule
        \textbf{\# Frames} & \textbf{Retrieval mean~(ms)} & \textbf{Packing mean/p95~(s)} & \textbf{Peak GPU Memory~(GiB)} \\
        \midrule
        64 & 33.2 & 1.139 / 1.289 & 18.875 \\
        128 & 33.3 & 1.080 / 1.203 & 18.880 \\
        256 & 33.9 & 1.142 / 1.283 & 18.878 \\
        512 & 34.0 & 1.144 / 1.239 & 18.890 \\
        1024 & 32.7 & 1.164 / 1.247 & 18.893 \\
        \bottomrule
    \end{tabular}
    \label{tab:detailed-profiling}
\end{table}

Our processing pipeline consists of three components:

\begin{enumerate}
    \item A CPU producer receives sampled frames and performs decoding and PNG encoding.
    \item A dedicated (foreground) GPU performs visual-token extraction and pooling.
    \item A second (background) GPU performs the text summarization required by gist memory and executes 
    consolidation when the memory budget is reached.
\end{enumerate}

Concretely, when each batch arrives, the coordinator first copies the frames 
or summarization and submits a background summarization job, while the foreground GPU performs 
visual extraction on the original images. After both sides finish, the coordinator constructs 
the gist entries in the original chunk order. When the budget is exceeded, the coordinator remains 
responsible for pair selection and sends only the two parent summaries together with a dynamically 
determined token cap to the background worker. Once the generated result is returned, the 
coordinator performs the visual merge, updates lineage metadata, conducts auditing, and commits 
the updated memory budget. The profiling results are shown at Table~\ref{tab:writer-profiling} are
collected with the same videos selected in Section~\ref{sec:exp} with 512-frame prefixes on A100 80G GPU.
It demonstrates that \sys could work at 1FPS. For deployment on edge devices, further 
architectural and computational optimization would still be required to accommodate the specific hardware platform.

\begin{table}[H]
    \centering
    \caption{Writer-side profiling metrics on subset of StreamingBench.}
    \begin{tabular}{lccc}
        \toprule
        \textbf{Metric} & \textbf{Mean} & \textbf{Median} & \textbf{p95} \\
        \midrule
        \textbf{Online (ms/frame)} & 911.35 & 937.39 & 1061.25 \\
        \textbf{Sustainable FPS} & 1.124 & 1.067 & 1.483 \\
        \textbf{Streaming update (ms/frame)} & 571.07 & 568.52 & 640.43 \\
        \textbf{Post-vision memory (ms/frame)} & 472.87 & 483.46 & 543.54 \\
        \bottomrule
    \end{tabular}
    \label{tab:writer-profiling}
\end{table}

\section{Case Study}
\label{app:case-study}
In this section, we'll use several examples to illustrate how the different 
memory roles in \sys work with each other.

\paragraph{Prompt Template.}
The prompt template we use is shown in Figure~\ref{fig:prompt}. Overall, it is also organized
into three parts corresponding to recent memory, episodic evidence, and gist 
memory. We adopt a timestamp-aware format: the query arrival time and each memory
segment are annotated with their corresponding timestamps, while gist entries 
are labeled with both start and end times to help the model establish temporal 
correspondences. We also include additional instructions on how the information 
provided by the three memory roles should be interpreted and integrated.
We find that, in some cases, the model mistakenly treats coarse summaries in gist
entries as key evidence for questions that require precise information, causing 
gist memory to become misleading instead. In addition, since recent frames may 
overlap with retrieved episodic evidence, we explicitly check for duplicates 
and avoid providing the same visual evidence to the model more than once, 
thereby reducing computational overhead and context usage.

\begin{figure}[H]
    \centering
    \begin{tcolorbox}[title=Prompt Template,fontupper=\small]

Use only the provided information about a video to answer the multiple-choice question.\br
The current time is \texttt{\{time:.3f\}} s. \br
Frames from recent 4 seconds: \br
\texttt{[recent frame images]}
\tcbline
Relevant Episode Evidence: \br
1. Frame at \texttt{\{time:.3f\}} s: \texttt{[image]} \br
...
\tcbline
Relevant Gist Memory: \br
1. Gist from \texttt{\{start:.3f\}} s to \texttt{\{end:.3f\}} s: \texttt{<text>} \br
...
\tcbline
Note: Recent frames and episode evidence should be the primary source for precise visual judgements. Gist memory is a lossy, high-level summary of
video segments and may omit or compress exact visual details, such as counts, colors, text, spatial positions, and fine-grained actions. If gist memory
conflicts with direct visual evidence, prefer the direct one. \br
Question: \texttt{question} \br
A. \texttt{<choice\_1>} \\
B. \texttt{<choice\_2>} \\
... \br
Answer with exactly one uppercase letter: A, B, C, D, or E.
    \end{tcolorbox}
    \caption{Prompt template used in \sys.}
    \label{fig:prompt}
\end{figure}

\begin{figure}[H]
    \centering
    \includegraphics[width=0.98\linewidth]{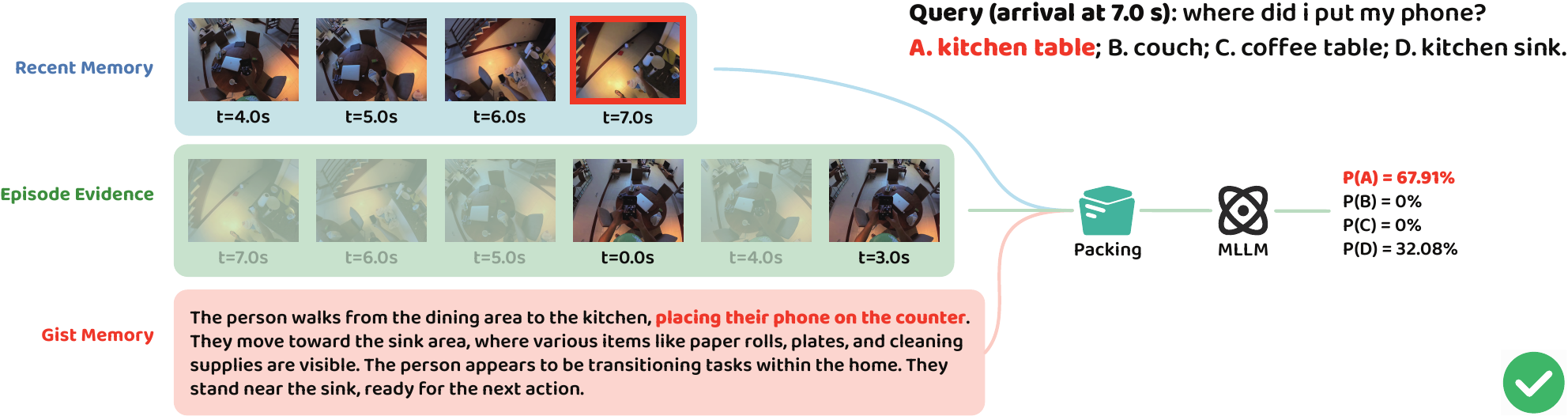}
    \caption{Case 1. The integration of R/E/G make the MLLM successfully get the correct answer.}
    \label{fig:case1}
\end{figure}

\paragraph{Case 1.}
Figure~\ref{fig:case1} presents an example in which the R/E/G memory roles
of \sys successfully enable the MLLM to produce the correct answer. The query 
arrives at (t=7.0) s and asks where the person in the first-person video places 
the phone. Recent memory consists of the four most recent frames, while episodic 
evidence ranks similar frames using a top-(k) retrieval strategy. Note that frames 
4, 5, 6, and 7 are already included in recent memory and are therefore not added 
again. Since the video segment is short, the single gist entry is directly included. 
These memory roles are then organized using the prompt template described above 
and provided to the MLLM. As shown, the frame at (t=7.0) s contains the key visual 
evidence that the phone is placed on the kitchen table, while the gist entry also 
explicitly states that the person places the phone on the counter. With these 
complementary cues, the model successfully selects the correct answer.

\begin{figure}[H]
    \centering
    \includegraphics[width=0.98\linewidth]{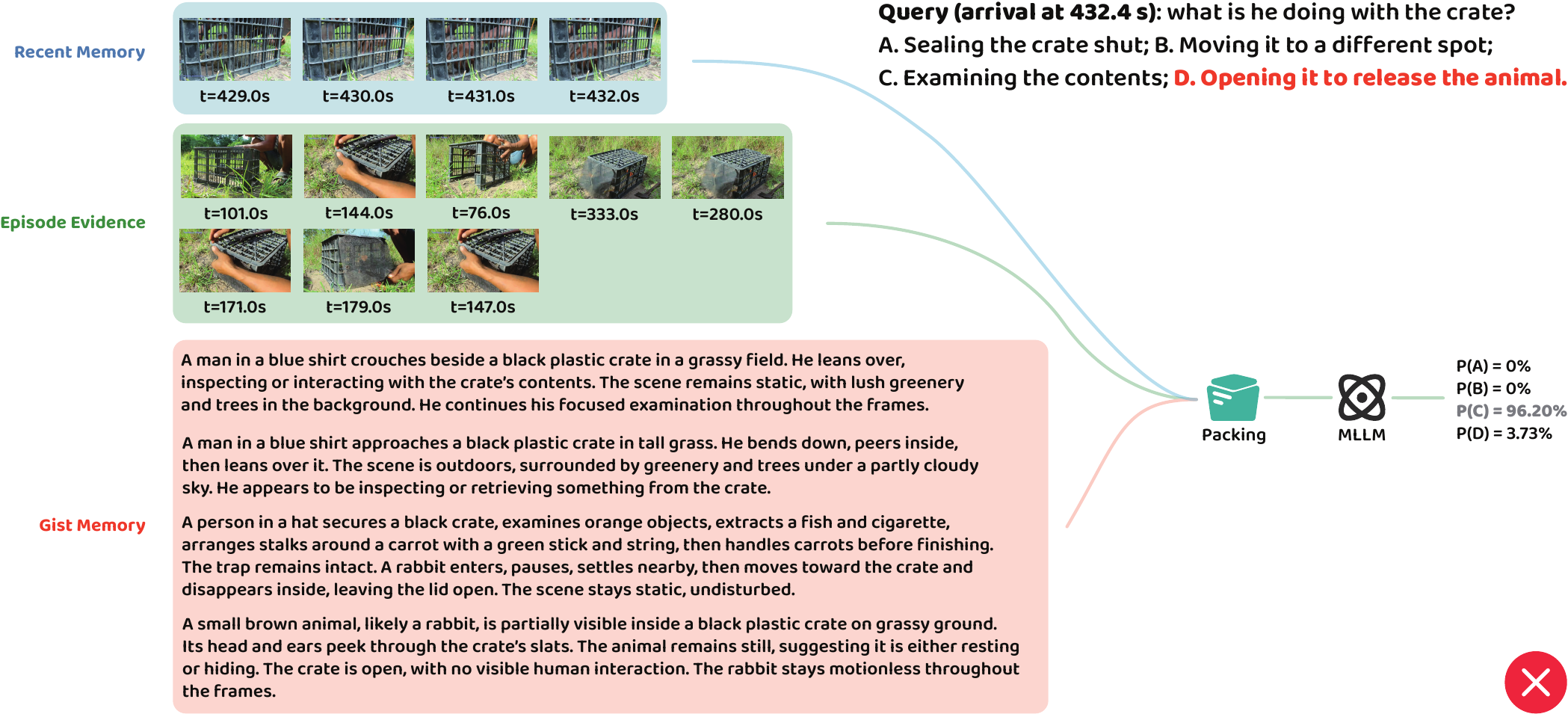}
    \caption{Case 2. The MLLM is confused with the time order of events and make a wrong decision.}
    \label{fig:case2}
\end{figure}

\paragraph{Case 2.}
Figure~\ref{fig:case2} presents a failure case in which the MLLM becomes confused
about the temporal order of events and ultimately produces an incorrect answer. 
The query arrives at \(t=432.4\) s and asks what the protagonist is doing to the 
crate \textit{at that moment}. The video is a tutorial in which the protagonist constructs 
a simple animal trap outdoors, a rabbit subsequently enters the cage, and the 
protagonist then prepares to release it. For this query, the correct answer can be 
inferred from recent memory: at that moment, the protagonist is releasing the rabbit. 
However, retrieval-based episodic evidence and gist memory are unable to properly model 
the temporal relationship, causing them to retrieve many events that occurred earlier 
in the video. This additional context misleads the model, which ultimately selects 
the wrong answer.

This example also highlights several limitations of the current \sys. First, the retrieval 
process does not explicitly account for temporal relationships. Second, answer prediction 
currently relies directly on the logits produced by a single model forward pass. Allowing 
the model to perform autoregressive reasoning could potentially improve final performance 
through test-time scaling. However, this introduces an additional context–performance 
trade-off, which we leave for future work.

\end{document}